\documentclass{article}

\usepackage{microtype}
\usepackage{graphicx}
\usepackage{xcolor}
\usepackage{booktabs} 
\usepackage{caption}
\usepackage{amsmath}
\usepackage{amssymb}
\usepackage{multirow}
\usepackage{verbatim}
\usepackage{epsfig}
\usepackage{subfig}
\usepackage{caption}

\usepackage{hyperref}

\newcommand{\ie}{\textit{i}.\textit{e}.}
\newcommand{\eg}{\textit{e}.\textit{g}.}

\newcommand{\Eg}{\textit{E}.\textit{g}.}

\usepackage[accepted]{icml2021}

\icmltitlerunning{Unsupervised Disentanglement without Autoencoding}

\begin{document}

\twocolumn[
\icmltitle{Unsupervised Disentanglement without Autoencoding: \\ Pitfalls and Future Directions}



\icmlsetsymbol{intern}{*}

\begin{icmlauthorlist}
\icmlauthor{Andrea Burns}{bu,goo}
\icmlauthor{Aaron Sarna}{goo}
\icmlauthor{Dilip Krishnan}{goo}
\icmlauthor{Aaron Maschinot}{goo}
\end{icmlauthorlist}

\icmlaffiliation{bu}{Department of Computer Science, Boston University, Boston, MA, USA}
\icmlaffiliation{goo}{Google Research, Cambridge, MA, USA}

\icmlcorrespondingauthor{Andrea Burns}{aburns4@bu.edu}

\icmlkeywords{Machine Learning, ICML}

\vskip 0.3in
]




\begin{abstract}
Disentangled visual representations have largely been studied with generative models such as Variational AutoEncoders (VAEs). While prior work has focused on generative methods for disentangled representation learning, these approaches do not scale to large datasets due to current limitations of generative models. Instead, we explore regularization methods with contrastive learning, which could result in disentangled representations that are powerful enough for large scale datasets and downstream applications.
However, we find that unsupervised disentanglement is difficult to achieve due to optimization and initialization sensitivity, with trade-offs in task performance.
We evaluate disentanglement with downstream tasks, analyze the benefits and disadvantages of each regularization used, and discuss future directions. 

\end{abstract}
\section{Introduction}
\label{sec:intro}
Learning semantically interpretable image representations has many benefits, such as better generalization \cite{eastwood2018a}, robustness to noise \cite{lopez2018information}, and increased transfer performance \cite{vansteenkiste2020disentangled}. Disentanglement can add interpretability if each representation dimension maps to one factor of variation present in the images while being relatively invariant to changes in the other factors \cite{bengio2014representation}; this is the definition of disentanglement that we adopt in this work.

Most unsupervised disentanglement approaches are built upon Variational AutoEncoders (VAEs) \cite{kingma2014autoencoding}, which are a type of generative model trained with a reconstruction and regularization loss. The regularization pushes the posterior distribution of the generative factors toward an isotropic Gaussian, encouraging its representations toward a factorized prior with independent dimensions, under the hope that this independence is close to the desired semantic disentanglement. However, because the reconstruction loss limits the amount of disentanglement that can be imposed in a vanilla VAE, these approaches have largely focused on augmenting the regularization loss to increase disentanglement without degrading reconstruction quality \cite{higgins2016beta,kim2019disentangling}. Generative Adversarial Networks (GANs) \cite{goodfellow2014generative} have also been used for disentanglement: the generator inputs are divided into latent and noise components, and a regularization loss is added to maximize the mutual information (MI) between the observations and latent components to the discriminator loss \cite{chen2016infogan,gulrajani2017improved}.

However, the losses that such models use to generate realistic images compete with the disentanglement regularization loss \cite{kim2019disentangling}
and the latent representations resulting from VAEs and GANs (disentangled or not) have significantly worse performance on downstream tasks such as retrieval and transfer learning \cite{pathak2016context, brock2018large}. We therefore consider disentanglement through contrastive representation learning \cite{chen2020simple,tian2020contrastive,oord2019representation}, with the hope of overcoming these limitations. Self-supervised contrastive learning frameworks achieve state-of-the-art performance for many tasks such as classification, detection and segmentation~\cite{chen2020simple,detection, segmentation}, so a disentangled contrastive representation that maintains task performance would be of great interest.

In the self-supervised setting, the contrastive loss is applied to the outputs of a shallow MLP (the `projection network') that takes the encoder output as input~\cite{chen2020simple}. Our initial hypothesis is that it may be feasible to disentangle
the encoder outputs 
while training the projection network. We explore three methods for disentanglement
with an unsupervised contrastive loss: 
minimizing MI, orthogonalizing to encourage independence, and reducing dependencies via Hessian regularization on the contrastive loss. While each regularizer has supporting intuition, in practice we find that issues arise due to disentanglement on undesired axes, initialization sensitivity, or optimization. This prevents the regularizers from reliably scaling to real-world datasets. 
We hope our results can be used to guide the community at large and spur new research directions.





\section{Contrastive Setup}
\label{sec:contrastive}
We use the SimCLR setup~\cite{chen2020simple} to study unsupervised disentanglement: encoder outputs $r_i$ are fed into a projection network and the contrastive loss is computed on projection outputs $z_i$ for $i$ in a batch of size $N$. 
We evaluate disentanglement 
on subembeddings $r_{i,k}$ or $z_{i,k}$, where $k\in[K]$ is a slice of $r_i$ or $z_i$. 
The InfoNCE loss~\cite{oord2019representation}
, which we refer to as InfoMax, is:
\begin{equation}
    \mathcal{L_{\text{InfoMax}}} = -\sum_{i \in I} \log{\frac{\exp({z_i \cdot z_{j} / \tau })}{\sum\limits_{a\in A}\exp({z_i \cdot z_{a} / \tau })}}
    \label{eq:infomax}
\end{equation}
where we have anchor $z_i$ and want to maximize MI with its positive $z_j$. Positives of the image $x_i$ are augmentations (\ie, `views') obtained via scaling, flipping, and color distortion transformations. The set $A$ includes the positive as well as all negatives in the batch with respect to the anchor.

We aim to disentangle subembeddings (slices of the representation) such that each subembedding encodes a distinct factor of variation. To do this, we maximize  MI between views of the same \textit{subembedding}. We adapt Eq. (\ref{eq:infomax}) to:
\begin{equation}
    \mathcal{L_{\text{SubInfoMax}}} = \sum_{k \in  K}\left(-\sum_{i \in I} \log{\frac{\exp({z_{i,k} \cdot z_{j,k} / \tau })}{\sum\limits_{a\in A}\exp({z_{i,k} \cdot z_{a,k} / \tau })}}\right)
\end{equation}
where $z_{j,k}$ is the $k$\textsuperscript{th} subembedding of the positive $z_j$ for anchor $z_i$. This results in $K$ InfoMax terms, which we sum over. Our experiments use the base case of two subembeddings (\ie, $K = 2$).
The regularization methods are added to the InfoMax objective as such for a regularizer $R$:
\begin{equation}
    \mathcal{L_{\text{Disentanglement}}} = \mathcal{L_{\text{SubInfoMax}}} + \lambda R 
    \label{eq:disentangle}
\end{equation}
where $\lambda$ weighs the regularization in the final objective.
\section{Experiment Setup}

\subsection{Dataset}
\label{sec:data}
We form a more realistic dataset for disentanglement, which we name MNIST/STL-10. It is similar to Colorful-Moving-MNIST~\cite{tian2020makes}, but designed for disentanglement. MNIST~\cite{lecun-mnisthandwrittendigit-2010} contains handwritten digits, and STL-10 is an image classification dataset~\cite{coates2011stl10}. 
We create MNIST/STL-10 instances by overlaying a MNIST digit on a STL-10 image. We can vary the following factors: digit class (DC), digit location (DL), or background class (BC). A view pair holds two of three factors constant while the third is varied uniformly at random\footnote{Examples can be found in the Appendix with additional dataset details. The dataset has been made publicly available at~\url{https://github.com/aburns4/UnsupDisent}.}


We run experiments with the DC-BC dataset. The dataset has two disjoint downstream tasks; digit (DC) and background (BC) classification. Upstream training fixes digit and background class in a view pair, but varies digit location. We train the encoder and projection network with joint labels, where each digit (ten classes) and background (ten classes) combination is one class, creating 100 classes. Separate downstream DC and BC linear classifiers are trained.


\subsection{Model Architecture}
\label{sec:arch}
We use a ResNet34~\cite{resnet} encoder, which outputs $r$, a 512D representation.
As compactness is a prerequisite for disentanglement (\ie, ~a factor of variation is represented with minimal elements) and our analysis finds our regularizers have a local effect on the representation (see Section~\ref{sec:infomin}), we also try using the projection output $z$ as input to the linear classifier $C$ used for our downstream tasks.




\subsection{Evaluation}
\label{sec:eval}
We compare accuracy when the full representation ($r$ or $z$) or subembeddings ($r_k$ or $z_k$) are used as classification input; subembeddings are halves of the full representation.
Table~\ref{tab:ideal_dcbc} shows the classification accuracy reflecting ideal disentanglement (achieved in a supervised setting, see details in Appendix).
Accuracy of $97.3\%$ and $64.5\%$ is achieved with $r$ for DC and BC, respectively and $r$ performs no better than either subembedding,
showing no loss of information. There also is a large gap $|C(r_0)-C(r_1)|$, where $r_1$ has near random performance on DC, and likewise for $r_0$ on BC. This diagonal accuracy trend suggests maximal disentanglement. 


\renewcommand{\arraystretch}{0.75}
\begin{table}[t]
    \centering
        \caption{Accuracy for digit (DC) and background (BC) classification tasks using a fully supervised pipeline with the full representation $r$ or its subembeddings $r_0$, $r_1$ as classification input.}
    \begin{tabular}{ccc}
    \toprule
    Classification Input & DC & BC \\
    \midrule
    $r$     & 97.3 & 64.5 \\
    \midrule
        $r_0$     & 97.3 & 10.1 \\
        \midrule
    $r_1$     & 11.7 & 64.5 \\
    \bottomrule
    \\
     $|C(r_0) - C(r_1)|$ & 85.6 & 54.4 
    \end{tabular}
    \vspace{-1mm}
    \label{tab:ideal_dcbc}
\end{table}
\section{Method}
\label{sec:method}
\subsection{InfoMin}
\label{sec:infomin}
Our first regularizer follows the intuition of wanting to \textit{minimize} MI between subembeddings, similar to reducing total correlation of latent codes used by~\citet{kim2019disentangling}.
We follow the approach of~\citet{tian2020makes} for view selection and define the first regularizer as $R_{\text{InfoMin}}=-\mathcal{L}_{\text{InfoMax}}$, but suspect a priori that it may cause optimization issues since a lower bound is now being minimized instead of maximized.
\begin{equation}
\label{infomin}
    R_{\text{InfoMin}} = \sum_{k\ne k'}\sum_{i \in I} \log{\frac{\exp({z_{i,k} \cdot z_{i,k'} / \tau })}{\sum\limits_{a\in A}\exp({z_{i,k} \cdot z_{a,k} / \tau })}}
\end{equation}
We let $k$ and $k'$ map to subembeddings of sample $z_i$ and sum over all such pairs. 
Eq. (\ref{infomin}) is computed over 
both views, as we compute the term $R_{\text{InfoMin}}$ within a single view. The regularization is applied to $z$ so that the InfoMin and InfoMax terms compete during the optimization of Eq. (\ref{eq:disentangle}). 

\subsubsection{InfoMin Experiments}
\label{sec:exps}



We try using $r$ or $z$ and their subembeddings as input to the downstream classifier. In the InfoMin pipeline, we use two projection heads, each with output dimension eight: $r_0$ and $r_1$ are passed to each network and result in $z_0$ and $z_1$. We define the full projection output $z$ as $z:=[z_0, z_1]$. 

Table~\ref{tab:infomin} summarizes disentanglement performance as measured by classification accuracy for InfoMin regularization. For perfect disentanglement, one would expect the DC accuracy for one of the two subembeddings (\eg, $r_0$ or $r_1$) to be equal to that achieved when using the full embedding ($r$), and the DC accuracy for the other subembedding would be equal to chance ($10\%$). For BC classification, the same would be true except that $r_0$ and $r_1$ would switch roles.

The results in Table~\ref{tab:infomin} are consistent with InfoMin not inducing disentanglement. The subembeddings have nearly the same downstream performance on each task; \eg, $z$, $z_0$, and $z_1$ have $\sim$74\% accuracy on DC when $\lambda=0.001$. We also see that InfoMin primarily affects the projection head output, demonstrating the regularization effects are local to the layer where it is applied. This can be seen as $r$, $r_0$, and $r_1$ perform nearly the same with different $\lambda$, while $z$, $z_0$, and $z_1$ performance drops heavily ($\sim$25\%) with $\lambda=0.1$.


We find $R_{\text{InfoMin}}$ has a direct trade-off with $\mathcal{L}_{\text{SubInfoMax}}$, potentially canceling out any disentanglement (see Appendix). 
As this approach causes optimization issues and hurts performance, we next explore approaches that enforce disentanglement without directly fighting the InfoMax objective. 

\begin{table}[h]
        \caption{Accuracy for digit (DC) and background (BC) classification tasks varying the weight $\lambda$ of the InfoMin term.}
    \centering
    \begin{tabular}{ccccc}
    \toprule
    \multirow{2}{*}{Classification Input} & \multicolumn{2}{c}{$\lambda=0.001$} & \multicolumn{2}{c}{$\lambda=0.1$} \\
    \cmidrule{2-5}
    & DC & BC & DC & BC \\
    \midrule
         $r$ & 97.0 & 63.1 & 96.4 & 61.1\\
         \midrule
         $r_0$ & 97.0 & 62.8 & 96.3 & 60.8 \\
         \midrule
         $r_1$ & 97.0 & 63.0 & 96.3 & 60.8\\
         \midrule
         $z$ & 74.3 & 60.7 & 47.6 & 35.7 \\
         \midrule
         $z_0$ & 74.4 & 60.6 & 47.5 & 35.7\\
         \midrule
         $z_1$ & 74.2 & 60.8 & 47.5 & 35.8\\
         \bottomrule
    \end{tabular}
    \label{tab:infomin}
\end{table}




\subsection{Orthogonality Constraint}
\label{sec:ortho}
Due to pitfalls experienced with $R_\text{InfoMin}$, we next 
tried constraining subembeddings to be orthogonal to enforce approximate (linear) independence. The resulting regularizer is an unsigned cosine distance between the subembeddings. Unlike $R_\text{InfoMin}$, computed between subembeddings from a single sample, we compute $R_\text{Ortho}$ between a subembedding $z_{i,k}$ of representation $z_i$ with subembeddings $z_{i,k'}$ \emph{and} $z_{j,k'}$
\begin{equation}
\label{ortho}
    R_{\text{Ortho}} =  \sum_{k\ne k'}\sum_i\sum_j\frac{|z_{i,k} \cdot z_{j,k'}|}{\|z_{i,k}\|\|z_{j,k'}\|}
\end{equation}
$R_{\text{Ortho}}$ is applied to $z$ due to the locality seen with $R_{\text{InfoMin}}$ and to have a lower dimension representation. 

\smallskip
\noindent\textbf{Degenerate Solutions} 
Several degenerate solutions may arise when minimizing $R_{\text{Ortho}}$. \Eg, two subembeddings could be rotations of one another, or the regularization could push one subembedding to a random vector. Both would satisfy Eq. (\ref{ortho}) without enforcing meaningful disentanglement. We found evidence of degenerate solutions experimentally as well, which are included in the Appendix. 

To address these issues, we also try adding a permutation matrix $P$ which randomly permutes one subembedding before computing the constraint; this should prevent degenerate solutions which rely on $z$'s element ordering to avoid penalty. Without loss of generality we permute subembedding $k$:
\begin{equation}
    R_{P(\text{Ortho})} =  \sum_{k\ne k'}\sum_i\sum_j\frac{|P(z_{i,k}) \cdot z_{j,k'}|}{\|z_{i,k}\|\|z_{j,k'}\|}
\end{equation}

\subsubsection{Orthogonality Experiments}
\label{sec:orthoexps}



\begin{table*}[t]
    \caption{Accuracy for digit (DC) and background (BC) classification tasks with orthogonality and Hessian regularization, varying the weight $\lambda$. For orthogonality regularization, ablations with the permutation mechanism ($P$) are also included. 
    } \centering
    \begin{tabular}{ccccccccccccc}
    \toprule
         \multirow{3}{*}{Classification Input} & \multicolumn{4}{c}{Orthogonality w/o $P$} & \multicolumn{4}{c}{Orthogonality w/ $P$} &\multicolumn{4}{c}{Hessian}\\
    \cmidrule{2-13}
    & \multicolumn{2}{c}{$\lambda=0.001$} & \multicolumn{2}{c}{$\lambda=0.1$} & \multicolumn{2}{c}{$\lambda=0.001$} & \multicolumn{2}{c}{$\lambda=0.1$}& \multicolumn{2}{c}{$\lambda=0.001$} & \multicolumn{2}{c}{$\lambda=0.1$}\\
    \cmidrule{2-13}
     & DC & BC & DC & BC & DC & BC & DC & BC & DC & BC & DC & BC \\
    \midrule
         $z$ & 97.0 & 62.4 & 69.3 & 59.6 & 97.0 & 62.2 & 97.2 & 61.6 & 92.4 & 57.2 & 93.6 & 56.3  \\
         \midrule
         $z_0$ & 76.6 & 48.4 & 51.2 & 39.9 & 77.1 & 52.4 & 88.0 & 56.9 & 60.7 & 46.7 & 74.7 & 42.4 \\
         \midrule
         $z_1$ & 73.2 & 54.0 & 46.3 & 42.7 & 71.8 & 57.0 & 82.2 & 60.0 & 47.9 & 34.6 & 57.5 & 50.9\\
         \bottomrule 
         \\
         $|C(z_0) - C(z_1)|$ & 3.4 & 5.6 & 4.9 & 2.8 & 5.3 & 4.6 & 5.8 & 3.1 & 12.8 & 12.1 & 17.2 & 8.5
    \end{tabular}
    \label{tab:ortho_hessian}
    \vspace{-2mm}
\end{table*}

The orthogonality contrastive pipeline also uses two projection heads (as described in Section~\ref{sec:exps}). In Table~\ref{tab:ortho_hessian}, we now see some disentanglement with $R_{\text{Ortho}}$, as nonzero differences (\eg, $|C_{DC}(z_0)-C_{DC}(z_1)|$, $|C_{BC}(z_0)-C_{BC}(z_1)|$, where $C(\cdot)$ denotes classification accuracy) are achieved.
Without permutation matrix $P$, the higher weight $\lambda=0.1$ hurts absolute performance using $z$ and its subembeddings. 
With $P$, performance is no longer hurt, and downstream subembedding performance increases with both $\lambda$ weights. This may be due to $P$ preventing subembedding slices from becoming random or rotations of other elements.

While the higher $\lambda$ does not increase $|C(z_0)-C(z_1)|$ values when $P$ is used, it does result in the highest subembedding performance over all experiments (88.0\% and 60.0\% for DC and BC tasks, respectively). Interestingly, the permutation $P$ results in a larger average subembedding difference for both tasks, having more impact than $\lambda$.
 

Using $R_{P(\text{Ortho})}$ with $\lambda=0.1$ shows the desired diagonal trend between $z_0$ and $z_1$, where the former performs better on DC (88.0 vs. 82.2) and the latter better on BC (60.0 vs. 56.9). However, the performance difference $|C(z_0) - C(z_1)|$ varied over multiple runs, with some having smaller values. Disentanglement also has higher variance on the BC task, and we include additional analysis in the Appendix. 

Ultimately, $R_{P(\text{Ortho})}$ resulted in some disentanglement, unlike $R_{\text{InfoMin}}$, but $|C(z_0)-C(z_1)|$ was not as large as the difference demonstrated with supervision in Table~\ref{tab:ideal_dcbc}. The permutation matrix also prevented loss of information, as seen by the improvement in performance across $z$ and its subembeddings. We'd like to retain downstream task performance while creating larger degrees of disentanglement. 
\subsection{Hessian Regularization}
\label{sec:hessian}
Although $R_{\text{Ortho}}$ avoids direct optimization trade-off with $\mathcal{L}_{\text{SubInfoMax}}$, it can still result in trivial solutions with
no disentanglement. Moreover, $R_{\text{Ortho}}$ did not create a significant gap in subembedding accuracy. We next explore constraining
the Hessian of $\mathcal{L}_{\text{InfoMax}}$ with respect to $z$, as it gives a measure of dependence between individual subembedding elements. We expect the Hessian of $z_i$ and $z_j$ to be zero if they are disentangled; this element-wise constraint may reduce trivial solutions. Computing the Hessian is expensive, but we can use the Gauss-Newton approximation:
\begin{equation}
    H_{ij} = \frac{\partial^2 \mathcal{L}_{\text{InfoMax}}}{\partial z_{i} \partial z_{j}} 
    \approx \frac{\partial \mathcal{L}_{\text{InfoMax}}}{\partial z_{i}} \cdot \frac{\partial \mathcal{L}_{\text{InfoMax}}}{\partial z_{j}}
\end{equation}

This is similar to the approach used by~\citet{peebles2020hessian}, but they constrain all off-diagonal terms to disentangle individual latent codes. 
Instead, we disentangle subembeddings with $\frac{d}{2}$ elements (for a $d$ element representation), allowing for intra-subembedding dependencies; \ie, we minimize the sum of the off-diagonal \textit{block} terms.

One InfoMax and one projection network are used, as the Hessian is computed on a unified representation $z$; $z_0$ and $z_1$ are now its halves.
The final regularizer is the Frobenius norm of the off-diagonal Hessian blocks:
\begin{equation}
    R_{\text{Hess}} = \sqrt{\sum_{i\in [k]} \sum_{j \in [k']} \left|\frac{\partial \mathcal{L}_{\text{InfoMax}}}{\partial z_{i}}\cdot\frac{\partial \mathcal{L}_{\text{InfoMax}}}{\partial z_{j}}\right|^2}
\end{equation}


\subsubsection{Hessian Experiments}
The Hessian regularizer only creates the desired diagonal accuracy (\ie, where one subembedding performs better on DC, and the other better on BC) with the higher $\lambda=0.1$. Yet both $\lambda$ values suggest that $R_{\text{Hess}}$ is a stronger regularizer than $R_{\text{InfoMin}}$ or $R_{\text{Ortho}}$, as $|C(z_0) - C(z_1)|$ grows larger. 

The diagonal performance trend found with the larger $\lambda$ was consistent in four out of five randomized experiments, significantly better than the one of out five experiments in the case of the orthogonality constraint. While $|C(z_0) - C(z_1)|$ is higher on average using $R_{\text{Hess}}$, it also has higher variance.
We include reproducibility analysis in the Appendix.

The downstream accuracy is similar using $z$ with either $\lambda$ weight, but $\lambda=0.1$ results in subembeddings $z_0$ and $z_1$ with higher performance (74.7 and 50.9 vs. 60.7 and 46.7 for the DC and BC tasks). Despite this, the strength of the Hessian constraint costs downstream accuracy, with lower accuracy using $z$, $z_0$, or $z_1$ compared to the $R_{\text{Ortho}}$ results.



In the end, $R_{\text{Hess}}$ is a strong enough constraint to enforce larger $|C(z_0)-C(z_1)|$, suggesting greater disentanglement, but it comes at the cost of downstream performance. The InfoMax loss may not balance this regularization enough, and the final objective may require additional losses to prevent loss of information in the representation. Using the Hessian and orthogonality terms together may also better balance the goal of disentanglement and high downstream accuracy.

\section{Discussion}
\label{sec:future}

Learning disentangled representations without autoencoding remains a difficult task with trade-offs in downstream accuracy, relative subembedding performance, and optimization. Classification accuracy suggests that enforcing orthogonality or reducing Hessian dependencies cannot achieve both high downstream performance and significant disentanglement.
Visualizing the representations learned via each regularizer may provide insight on how to balance both desired qualities. 
We also find a loss of information, as neither subembedding performs as well as the full representation; additional constraints enforcing the full representation performs no better than either subembedding may help to prevent this. Lastly, additional inductive biases may be needed for greater disentanglement without hurting task performance.


\bibliography{egbib}
\bibliographystyle{icml2021}
\newpage
\appendix
\section{Appendix}
\subsection{MNIST/STL-10 Dataset}
\subsubsection{Dataset Creation}

\begin{figure}[h]
\centering
\includegraphics[width=0.4\textwidth]{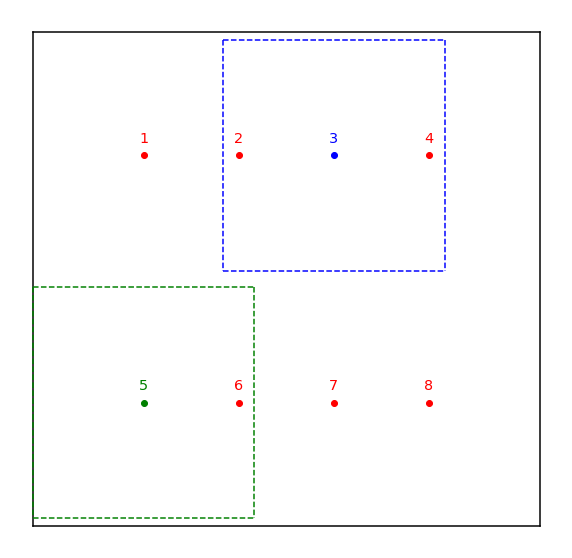}
\caption{Possible digit locations (DLs) at which to center the MNIST instance on a STL-10 image. Each MNIST instance is 28x28, and each STL-10 instance is 64x64. Digit locations were chosen to be uniformly distributed in 2 rows and 4 columns such that the entire MNIST instance was always overlaid. MNIST bounding box examples are shown for location 3 (in blue) and 5 (in green).}
\label{fig:digit_locations}
\end{figure}

We create MNIST/STL-10 by overlaying MNIST~\cite{lecun-mnisthandwrittendigit-2010} handwritten digits on STL-10~\cite{coates2011stl10} images. Each overlayed MNIST image is centered at one of eight possible pixel locations in the STL-10 image (Figure~\ref{fig:digit_locations}). Since MNIST digits tend to saturate each color channel, a simple per-pixel maximum of the MNIST digit image and the STL-10 image was used. An MNIST/STL-10 instance is generated by specifying three properties: 1) the MNIST digit instance belonging to one of 10 classes (0 - 9); 2) the STL-10 instance belonging to one of 10 classes (airplane, bird, car, cat, deer, dog, horse, monkey, ship, truck); and 3) the pixel location at which to center the MNIST instance on the STL-10 instance (8 possible values).

As we have three factors of variation (digit class (DC), background class (BC), and digit location (DL)), we can create three dataset variants: DC-BC, DC-DL, and BC-DL. In each of these MNIST/STL-10 variants, we hold two of the three factors of variation constant in a view pair. So, DC-BC holds digit class and background class constant in view pairs while digit location is varied, DC-DL holds digit class and digit location constant in view pairs while background class is varied, and so on. Each dataset variant has a total of 100k train samples and 10k test samples.

Note that when we fix a factor of variation, we do not fix a particular instance in that view pair. \Eg, if a view pair of the DC-BC dataset has DC set to the digit one, each view can contain a different instance of the one class, as seen in Figure~\ref{fig:dcbc}.

\subsubsection{Dataset Examples}
\begin{figure}[h]
\centering
\subfloat[View 0 in view pair]{\label{fig:mdleft}{\includegraphics[width=0.24\textwidth]{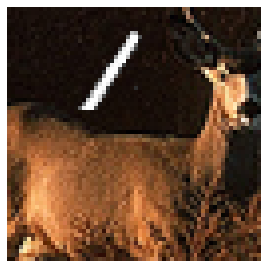}}} \hfill
\subfloat[View 1 in view pair]{\label{fig:mdright}{\includegraphics[width=0.24\textwidth]{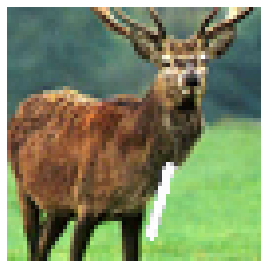}}}
\caption{Dataset examples from the MNIST/STL-10 DC-BC variant.
}
\label{fig:dcbc}
\end{figure}

We include view pairs from all three dataset variants to illustrate what the MNIST/STL-10 samples look like. In Figure~\ref{fig:dcbc}, we show a view pair from the DC-BC dataset. In this dataset, the digit location is varied between views. View 0 lays the digit in the upper middle-left position (location 2), while view 1 lays the digit in the lower middle-right position (location 7). Fixed between views is both the digit class (one) and background class (deer). As seen in the view pair, the instances of the written digit and image class are varied.

\begin{figure}[h]
\centering
\subfloat[View 0 in view pair]{\label{fig:mdleft}{\includegraphics[width=0.24\textwidth]{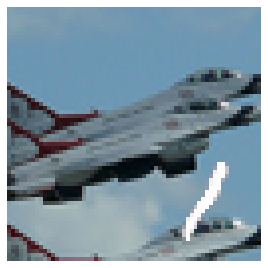}}} \hfill
\subfloat[View 1 in view pair]{\label{fig:mdright}{\includegraphics[width=0.24\textwidth]{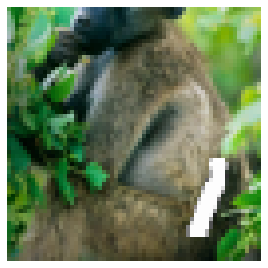}}}
\caption{Dataset examples from the MNIST/STL-10 DC-DL variant.}
\label{fig:dcdl}
\end{figure}

\begin{figure}[h]
\centering
\subfloat[View 0 in view pair]{\label{fig:mdleft}{\includegraphics[width=0.24\textwidth]{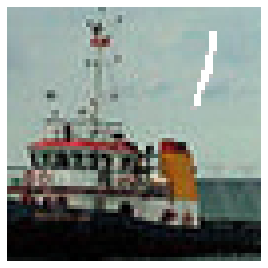}}} \hfill
\subfloat[View 1 in view pair]{\label{fig:mdright}{\includegraphics[width=0.24\textwidth]{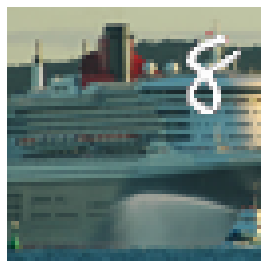}}}
\caption{Dataset examples from the MNIST/STL-10 BC-DL variant.}
\label{fig:bcdl}
\end{figure}

Figure~\ref{fig:dcdl} and Figure~\ref{fig:bcdl} include view pair examples from the DC-DL and BC-DL datasets. In a DC-DL view pair the digit class and digit location are fixed, as can be seen by the digit one and lower right digit position (location 8) held constant in the view pair of Figure~\ref{fig:dcdl}. The background class in this case is varied, with view 0 containing the plane class, while view 1 contains the monkey class. Lastly, Figure~\ref{fig:bcdl} illustrates a view pair from the BC-DL MNIST/STL-10 dataset variant, in which digit class is varied between views and background class and digit location are held constant. This can be seen as view 0 contains the digit one, while view 1 contains the digit eight, but both are located in the upper right most position (location 4) atop images of ships.

\subsection{Fully Supervised Pipeline}
\label{sec:sup}
While we are interested in a fully unsupervised contrastive learning pipeline for learning disentangled representations, we verify the feasibility of our general approach via a fully supervised experiment setting. 

We split $r$ into its halves $r_0$ and $r_1$, and apply the InfoMax loss to each half using \textit{disjoint} labels. Instead of using joint labels as described in Section~\ref{sec:data}, we separate the task labels for each InfoMax. \Eg, for the DC-BC dataset, $\mathcal{L}_{\text{SubInfoMax}}$ is applied to $r_0$ using the DC labels (ten MNIST digit classes) to define positive and negative samples, and $L_{\text{SubInfoMax}}$ is applied to $r_1$ using the BC labels (ten STL-10 image classes) to define positive and negative samples.

\subsection{InfoMin-Max Trade-Off}
\begin{figure*}[h]
\centering
\subfloat[The InfoMax of subembedding 0]{\label{fig:mdleft}{\includegraphics[width=0.28\textwidth]{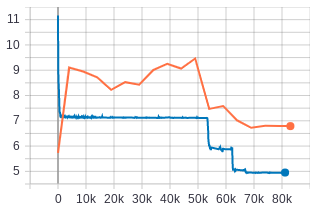}}}\hfill
\subfloat[The InfoMax of subembedding 1]{\label{fig:mdright}{\includegraphics[width=0.3\textwidth]{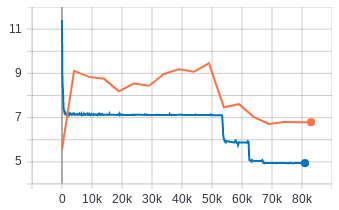}}}\hfill
\subfloat[The InfoMin between subembeddings]{\label{fig:mdright}{\includegraphics[width=0.3\textwidth]{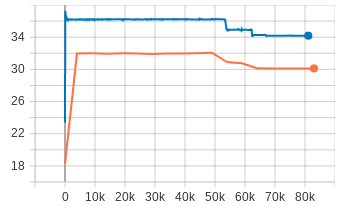}}}

\caption{Losses during upstream training of the DC-BC dataset with InfoMin. We want the InfoMax loss to decrease (maximizing MI between views) and the InfoMin to increase (minimizing MI between subembeddings). Near 55k steps a trade-off occurs, in which minimizing the InfoMax is prioritized. This results in greater subembedding entanglement, as seen by the undesired drop in the InfoMin.}
\label{fig:infomin}
\end{figure*}

We suspected that there is a direct trade-off between the InfoMin and InfoMax loss terms, which is confirmed by the loss visualization provided in Figure~\ref{fig:infomin}. The optimization is stagnant until nearly 55k steps, at which point the InfoMax terms `win' and the InfoMin values drop. 
Unfortunately, the higher weighted InfoMin term with $\lambda=0.1$ does not resolve this trade-off and only hurts the optimization of the InfoMax, reducing the representation quality.

\subsection{DC-DL Experiments}
The main paper includes experiments on the DC-BC dataset, and we report additional results on the DC-DL dataset here due to space. 
\subsubsection{Ideal Disentanglement}
Similar to the accuracies reported in Table~\ref{tab:ideal_dcbc}, we report the results on DC-DL when training the contrastive pipeline using the fully supervised setup described in Section~\ref{sec:sup}. Table~\ref{tab:ideal_dcdl} shows that $r_0$ performs well on the DC task, specifically just as well as $r$; the same holds for $r_1$ and the DL downstream task. The resulting performance differences between subembeddings show maximal accuracy differences, with each subembedding performing near random for the other task.
 
\begin{table}[h]
    \centering
        \caption{Accuracy for digit class (DC) and digit location (DL) classification tasks using a fully supervised pipeline with the full representation $r$ or its subembeddings $r_0$, $r_1$ as classification input.}
    \begin{tabular}{ccc}
    \toprule
    Classification Input & DC & DL \\
    \toprule
    $r$     & 98.2 & 99.9 \\
    \midrule
    $r_0$     & 98.2 & 13.4 \\
    \midrule
    $r_1$     & 11.7 & 99.9 \\
    \bottomrule
    \\
     $|C(r_0) - C(r_1)|$ & 86.5 & 86.5
    \end{tabular}
    \label{tab:ideal_dcdl}
\end{table}

\subsubsection{InfoMin}
\begin{table}[]
    \caption{Accuracy for digit class (DC) and digit location (DL) classification tasks using the InfoMin term.}
    \centering
    \begin{tabular}{ccc}
    \toprule
    Classification Input & DC & DL \\
    \midrule
        $r$ & 11.7 & 12.6  \\
        \bottomrule
    \end{tabular}
    \label{tab:infomin_dcdl}
\end{table}
Using the InfoMin regularization is ineffective with the DC-DL MNIST/STL-10 dataset variant. As seen in Table~\ref{tab:infomin_dcdl}, performance on both downstream tasks is near random. This is due to the InfoMin directly combatting the InfoMax terms in the final objective, which prevents optimization. As a result, no meaningful information is encoded in the representation.
\subsubsection{Orthogonality}
Table~\ref{tab:ortho_hessian_dcdl} includes experiment results using the orthogonality constraint with and without the permutation matrix $P$ and using various $\lambda$ weights. While there are subembedding performance gaps ($|C(z_0)-C(z_1)|$) on the DC task without the permutation $P$ being used, there is little difference between the subembeddings' performance for the DL task. As the orthogonality regularizer is not an element-wise constraint, digit location information may be easily repeated within a subembedding and may not be penalized by this regularizer. The $|C_{DL}(z_0)-C_{DL}(z_1)|$ difference does increase with the higher $\lambda=0.1$ (0.1 vs. 2.8).

Surprisingly, the performance differences are smaller when the permutation matrix is used. Additionally, the accuracy achieved when $z$ is used as classification input is lower with $R_{P(\text{Ortho})}$, which may be due to disentanglement being enforced on axes that are not related to the downstream tasks.
\begin{table*}[t]
    \caption{Accuracy for digit class (DC) and digit location (DL) classification tasks with orthogonality and Hessian regularization, varying the weight $\lambda$. For orthogonality regularization, ablations with the permutation matrix ($P$) are also included. }
    \centering
    \begin{tabular}{ccccccccccccc}
    \toprule
    \multirow{3}{*}{Classification Input} & \multicolumn{4}{c}{Orthogonality w/o P} & \multicolumn{4}{c}{Orthogonality w/P} & \multicolumn{4}{c}{Hessian}\\
    \cmidrule{2-13}
    & \multicolumn{2}{c}{$\lambda=0.001$} & \multicolumn{2}{c}{$\lambda=0.1$} & \multicolumn{2}{c}{$\lambda=0.001$} & \multicolumn{2}{c}{$\lambda=0.1$} & \multicolumn{2}{c}{$\lambda=0.001$} & \multicolumn{2}{c}{$\lambda=0.1$}\\
        \cmidrule{2-13}

    & DC & DL & DC & DL & DC & DL & DC & DL & DC & DL & DC & DL \\
    \midrule
        $z$ & 82.6 & 99.6 & 97.8 & 99.6 & 96.6 & 99.8 & 68.0 & 99.8 & 86.6 & 99.1 & 91.2 & 99.1 \\
            \midrule
        $z_0$ & 44.3 & 99.4 & 46.5 & 83.9 & 45.5 & 99.6 & 41.3 & 99.7 & 46.7 & 41.5 & 39.9 & 48.1\\
            \midrule
        $z_1$ & 33.6 & 99.5 & 38.4 & 86.7 & 44.5 & 99.7 & 34.5 & 98.6 & 41.7 & 67.9 & 34.8 & 64.7 \\
        \bottomrule
        \\
        $|C(z_0)-C(z_1)|$ & 10.7 & 0.1 & 8.1 & 2.8 & 1.0 & 0.1 & 6.8 & 1.1 & 5.0 & 26.4 & 5.1 & 16.6
    \end{tabular}
    \label{tab:ortho_hessian_dcdl}
\end{table*}

\subsubsection{Hessian}
Similar to the DC-BC results in Table~\ref{tab:ortho_hessian}, the Hessian constraint can result in larger performance differences between the two subembeddings. Notably, there are much larger differences for the digit location task, $|C_{DL}(z_0)-C_{DL}(z_1)|$. 
The Hessian regularizer is the first to break the same performance of $z_0$ and $z_1$ on the digit location task (\ie, all other regularization ablations in Table~\ref{tab:ortho_hessian_dcdl} have about $99\%$ accuracy using $z$ or its subembeddings as classification input).

\subsubsection{Degenerate Solutions}
We found the DC-DL dataset variant is more prone to extreme degenerate solutions, as illustrated in Table~\ref{tab:ortho_degen}. This may in part be due to a task difficulty imbalance: digit location information may be significantly more easily learned and encoded than digit class. Table~\ref{tab:ortho_degen} includes experiment results from using $R_\text{Ortho}$ with $\lambda=0.1$, which only provides a whole representation constraint instead of an element-wise constraint such as the Hessian.
The digit location information may be repeated multiple times within the full representation; this can be seen in the Task Difficulty columns of Table~\ref{tab:ortho_degen}, in which DL accuracy is constant across $z$, $z_0$, $z_1$.  


Another example of a trivial solution is shown in the first two columns of Table~\ref{tab:ortho_degen}, in which performance on both downstream tasks is near random. The orthogonality constraint is too strong and prevents optimization during training, resulting in random representations. The DC-DL dataset may be more prone to such optimization issues because of task difficulty imbalance, too.

\begin{table}[t]
    \centering
    \caption{Degenerate solutions on digit class (DC) and digit location (DL) classification tasks using the orthogonality term without the permutation matrix $P$ and a weight of $\lambda=0.1$.}
    \begin{tabular}{ccccc}
    \toprule
    \multirow{2}{*}{Classification Input} & \multicolumn{2}{c}{Random} & \multicolumn{2}{c}{Task Difficulty} \\
    \cmidrule{2-5}
    & DC & DL & DC & DL\\
    \midrule
         $z$ & 11.7 & 12.6 & 76.7 & 99.8\\
        \midrule
         $z_0$ & 11.7 & 12.6 & 42.0 & 99.7\\
         \midrule
         $z_1$ & 11.7 & 12.6 & 32.1 & 99.7\\
         \bottomrule
    \end{tabular}
    \label{tab:ortho_degen}
\end{table}


\subsection{Reproducibility}
\begin{table*}[h]
    \centering
    \caption{Reproducibility results on downstream digit (DC) and background (BC) classification tasks using the orthogonality regularizer with the permutation matrix $P$ and a weight of $\lambda=0.1$. Run 0 is the original result reported in Table~\ref{tab:ortho_hessian}, runs 1-5 are reruns that have different random initialization.}
    \begin{tabular}{ccccccccccccc}
    \toprule
    \multirow{2}{*}{Classification Input} & \multicolumn{2}{c}{Run 0} & \multicolumn{2}{c}{Run 1} & \multicolumn{2}{c}{Run 2} & \multicolumn{2}{c}{Run 3} & \multicolumn{2}{c}{Run 4} & \multicolumn{2}{c}{Run 5} \\
    \cmidrule{2-13} 
    & DC & BC & DC & BC & DC & BC & DC & BC & DC & BC & DC & BC\\
    \midrule
         $z$ & 97.2 & 61.6 & 97.0 & 62.3 & 97.0 & 61.3 & 97.2 & 61.6 & 97.2 & 61.4 & 97.0 & 58.3 \\
         \midrule
         $z_0$ & 88.0 & 56.9 &86.8 & 59.3 & 85.4 & 57.8 & 83.0 & 59.7 & 80.8 & 57.7 & 71.6 & 56.5 \\
         \midrule
         $z_1$ & 82.2 & 60.0 & 80.8 & 59.6 & 79.6 & 57.6 & 81.2 & 56.2 & 75.6 & 54.7 & 69.7 & 48.1 \\
         \bottomrule \\
         $|C(z_0) - C(z_1)|$ &5.8 & 3.1 & 6.0 & 0.3 & 5.8 & 0.2 & 1.8 & 3.5 & 5.2 & 3.0 & 1.9 & 8.4
         
    \end{tabular}
    \label{tab:ortho_reprod}
\end{table*}

\begin{table*}[t]
    \centering
    \caption{Reproducibility results on downstream digit (DC) and background (BC) classification tasks using the Hessian regularizer and a weight of $\lambda=0.1$. Run 0 is the original result reported in Table~\ref{tab:ortho_hessian}, runs 1-5 are reruns that have different random initialization.}
    \begin{tabular}{ccccccccccccccc}
    \toprule
    \multirow{2}{*}{Classification Input} & \multicolumn{2}{c}{Run 0} & \multicolumn{2}{c}{Run 1} & \multicolumn{2}{c}{Run 2} & \multicolumn{2}{c}{Run 3} & \multicolumn{2}{c}{Run 4} & \multicolumn{2}{c}{Run 5} \\
    \cmidrule{2-13} 
    & DC & BC & DC & BC & DC & BC & DC & BC & DC & BC & DC & BC\\
    \midrule
         $z$ & 93.6 & 56.3 & 73.0 & 56.0 & 96.2 & 56.6 & 93.9 & 56.6 & 73.1 & 58.1 & 92.2 & 56.9 \\
         \midrule
         $z_0$ & 74.7 & 42.4 & 67.9 & 47.9 & 73.7 & 40.4 & 58.4 & 36.6 & 42.2 & 34.8 & 61.1 & 36.4\\
         \midrule
         $z_1$ & 57.5 & 50.9 & 64.2 & 49.4 & 66.5 & 43.2 & 51.0 & 37.5 &  40.0 & 30.2 & 53.5 & 37.2
         \\
         \bottomrule \\
         $|C(z_0) - C(z_1)|$ & 17.2 & 8.5 & 3.7 & 1.5 & 7.2 & 2.8 & 7.4 & 0.9 & 2.2 & 4.6 & 7.6 & 0.8

    \end{tabular}
    \label{tab:hessian_reprod}
\end{table*}

We include five additional DC-BC experiments for both the orthogonality and Hessian constraints to see how performance varies over different random initializations. Experiments are rerun under the best settings for each regularizer: $R_{P(\text{Ortho})}$ and $R_{\text{Hess}}$, both with a weight of $\lambda=0.1$ in the final training objective.

\subsubsection{Orthogonality Constraint}
The first run in Table~\ref{tab:ortho_reprod}, run 0, is the original experiment included in the results of Table~\ref{tab:ortho_hessian}. The diagonal trend of one subembedding performing better on DC and one subembedding performing better on BC only occurs once in the additional five runs, with a barely nonzero difference $|C(z_0)-C(z_1)|$ for the background classification task (0.3). 

In the four experiments with no diagonal performance trend, the subembeddings $z_0$ and $z_1$ either perform nearly the same on background classification (\eg, run 2), or one subembeddings performs better on both tasks (\eg, run 3, 4, 5).

The mean difference in subembedding performance for the digit classification task, \eg, $\frac{1}{R}|C_{\text{DC}}(z_0)-C_{\text{DC}}(z_1)|$ over $R$ runs, is 4.4, and the differences $|C_{\text{DC}}(z_0)-C_{\text{DC}}(z_1)|$ have a variance of 3.4. For background classification the mean difference is 3.1, with a higher variance of 7.4. This demonstrates that the digit classification task is prioritized during optimization as it has larger performance gaps with lower variance. 

\subsubsection{Hessian Regularization}
The first run in Table~\ref{tab:hessian_reprod}, run 0, is the original experiment included in the results of Table~\ref{tab:ortho_hessian}. Among the additional five experiments, four satisfied the desired diagonal performance trend, with one subembedding performing best on the DC task, and the other performing best on the BC task.

The mean difference in subembedding performance for the digit classification task, \eg, $\frac{1}{R}|C_{\text{DC}}(z_0)-C_{\text{DC}}(z_1)|$ over $R$ runs, is 7.6, which is larger than the 4.4 mean difference obtained using the orthogonality constraint. However, the variance of performance differences ($|C_{\text{DC}}(z_0)-C_{\text{DC}}(z_1)|$) is significantly larger using the Hessian (22.8 variance of subembedding performance differences on the DC task vs. 3.4 variance when $R_{P(\text{Ortho})}$ is used). 

On the other hand, the mean subembedding performance difference increases and the variance of these differences decreases on the background classification task, showing different performance trends for each downstream task. The mean BC difference over all experiments, $\frac{1}{R}|C_{\text{BC}}(z_0)-C_{\text{BC}}(z_1)|$, is 3.2 and variance of $|C_{\text{BC}}(z_0)-C_{\text{BC}}(z_1)|$ is 7.4. This is negligibly better than the mean of 3.1 and variance of 7.4 obtained from the orthogonality results.

\end{document}